\documentclass{article}
\pdfoutput=1

\usepackage{authblk}

\usepackage[squaren]{SIunits}   \usepackage{courier}            \usepackage[scaled]{helvet}     \usepackage{url}                \usepackage{listings}           \usepackage{enumitem}           \usepackage[colorlinks=true,allcolors=blue,breaklinks,draft=false]{hyperref}      
\usepackage[pdftex]{graphicx}

\usepackage{amsthm}
\usepackage{amsmath}
\usepackage{amssymb}
\usepackage{dsfont}             \usepackage{xspace}
\usepackage{multirow}

 \theoremstyle{definition}                                     \newtheorem{definition}{Definition}
\theoremstyle{definition}                                     \newtheorem{example}{Example}                                                                                               
\newcommand{\cut}[1]{ }
\newcommand{\thetech}{{\sc Psi}\xspace}
\newcommand{\inter}[2]{[#1 \leftarrow #2]}
\newcommand{\cf}[3]{{#1}^{#2}_{#3}}
\newcommand{\cfR}[2]{{#1}_{#2}}
\newcommand{\irl}[1]{$\mathtt{#1}$}

\long\def\authornote#1{  \leavevmode\unskip\raisebox{-3.5pt}{\rlap{$\scriptstyle\diamond$}}  \marginpar{\raggedright\hbadness=10000
    \def\baselinestretch{0.8}\tiny
    \it #1\par}}

\newcommand{\aleks}[1]{\authornote{AC: #1}}

\begin{document}

\title{Debugging Machine Learning Tasks}

\author[A]{Aleksandar Chakarov}
\author[B]{Aditya Nori}
\author[B]{Sriram Rajamani}
\author[C]{\\Shayak Sen}
\author[D]{Deepak Vijaykeerthy}

\affil[A]{University of Colorado, Boulder}
\affil[B]{Microsoft Research}
\affil[C]{Carnegie Mellon University}
\affil[D]{IBM Research}

\maketitle
\begin{abstract}
Unlike traditional programs (such as operating systems or word processors) which have large amounts of code, machine learning tasks use programs with relatively small amounts
of code (written in machine learning libraries), but voluminous amounts of data. Just like developers of traditional programs debug errors in their code,
developers of machine learning tasks debug and fix errors in their data. However, algorithms and tools for debugging and fixing errors in data are
less common, when compared to their counterparts for detecting and fixing errors in code.
In this paper, we consider classification tasks where errors in training data lead to misclassifications in test points, and propose an automated
method to find the root causes of such misclassifications. Our root cause analysis is based on Pearl's theory of causation, and uses Pearl's PS (Probability of
Sufficiency) as a scoring metric. Our implementation, \thetech, encodes the computation of PS as a probabilistic program, and uses recent work
on probabilistic programs and transformations on probabilistic programs (along with gray-box models of machine learning algorithms)
to efficiently compute PS. \thetech is able to identify root causes of
data errors in interesting data sets.
\end{abstract}

\section{Introduction}
\label{Section:Introduction}
Machine learning techniques are used to perform data-driven decision-making in
a large number of diverse areas including  image processing,  medical
diagnosis, credit decisions, insurance decisions, email spam detection, speech
recognition, natural language processing, robotics, information retrieval and
online advertising.  Over time, these techniques have been honed and tuned, and
are now at a stage where machine learning libraries~\cite{weka,scikit-learn} are used as black-boxes by
programmers with little or no expertise in the details of the machine learning
algorithms themselves.  The black-box nature of the reuse, however, has an
unfortunate downside.  Current implementations of machine learning techniques
provide little insight into why a particular decision was made. Because of this absence
of transparency, debugging the outputs of a machine learning algorithm has
become incredibly hard.

Most programmers who implement machine learning use libraries to build models
from voluminous training data, and then use these models to perform
predictions.  These machine learning libraries often employ complex,
stochastic, or approximate, search and optimization algorithms that search for
an optimal model for a given training data set.  The model is then applied to a
set of unseen test samples in the hope of satisfactory generalization.  When
generalization fails, i.e., an incorrect result is produced for a test input, it
is often difficult to debug the cause of the failure.  Such failures can arise
due to several reasons.  Common causes for failure include bugs in the
implementation of the machine learning algorithm, incorrect choice of features,
incorrect setting of parameters (such as degree of the polynomial for
regression or number of layers in a neural network) when invoking the machine
learning library, and noise in the training set.  Over time, bugs in
implementation of machine learning algorithms get detected and fixed. There is
a lot of work in feature selection~\cite{feature-selection}, and
parameter choices can be made by systematically building models for various
parameter values and choosing the model with the best validation
score~\cite{parameter-search}.  However, since training data is typically
voluminous, errors in training data are common and notoriously difficult to
debug. This suggests a new class of debugging problems where programs (machine
learning classifiers) are learnt from data and bugs in a program are now the result 
of faults in the data.

In this paper, we focus on debugging machine learning tasks in the presence of
errors in training data.  Specifically we consider classification tasks, which
are typically implemented using algorithms such as logistic
regression~\cite{bishop} and boosted decision trees~\cite{Burges10}. Suppose we train
a classifier on training data (which has errors), and the classifier produces
incorrect results for one or more test points.  We desire to produce an
automated procedure to identify the \emph{root cause} of this failure. That is,
we would like to identify a subset of training points that influences the
classification for these test points the most. Therefore, correcting mistakes
in these training points is most likely to fix the incorrect results.

Our algorithm for identifying root causes is inspired by the structural
equations framework of causation, as formulated by Judea
Pearl~\cite{ProbCausation,CausalityBook}.  We think of each of the training
data points as possible causes of the misclassification in the test data set,
and calculate for each such training point, a score corresponding to how likely
it is that the current label for that point is the cause for the
misclassification of the test data set.  A simple measure of the score of a
training point can be obtained by merely flipping the label of the training
point and observing if the flip improves the results of the classifier on test
points. However, such a simple measure does not work when errors exist in
several training points, and several training points {\em together} cause the
incorrect results in the test points.  Thus, the score we calculate for each
training point $t$ considers alternate \emph{counterfactual worlds}, where training
points are labeled with several possible values (other than the value in the
training data), and sums up the probability that flipping the label of $t$
causes the misclassification error in the test data, among all such alternate
worlds. In Pearl's framework, such a score is called the {\em probability of
sufficiency} or PS for short.

One of the main difficulties in calculating the probability of sufficiency is 
that the classifier (or model) needs to be relearnt for alternate
worlds. Each of these model computing steps (also called as {\em training}
steps) is expensive.  We use a ``gray box'' view of the machine learning
library, and profile key intermediate values (that are hand-picked for each
machine learning algorithm) during the initial training phase. Using these
values, we build a gray-box abstraction of the training process by which the
model for a new training set (which is obtained by flipping certain number of
training labels) can be obtained efficiently without the need to perform
complete (and expensive) retraining.  Finally, we are able to amortize the cost
of computing the PS score by sharing common work {\em across} the computation
for different training points.

In order to carry out these optimizations, we model the PS computation as a
{\em probabilistic program}~\cite{fose}. Probabilistic programs
allow us to represent all of the above optimizations such as using gray-box
models, using instrumented values from actual training runs, and sharing work
across multiple PS computations as program transformations.  We are also able
to leverage recent progress in efficient inference of probabilistic programs to
scale the computation of PS scores to large data sets.

We have implemented our root cause detection algorithm in a tool \thetech.
\thetech currently works with two popular classifiers: (1) logistic regression,
and (2) boosted decision trees.  For these classifiers, \thetech runs a production
quality implementation of the techniques, profiles specific values and builds
an abstract gray-box model of the classifier, which avoids expensive
re-training.  Armed with this gray-box model, \thetech performs scalable inference
to compute the PS values for all points in the training set.  \thetech is able to
identify root causes of misclassifications in several interesting data sets.
In summary, the main contributions of this paper are as follows:
\begin{itemize}
\item We propose using the structural equations framework of causality, and specifically Pearl's PS
score to compute root causes of failures  in machine learning algorithms.
\item We model the PS computation as a probabilistic program, and this enables us to
leverage efficient techniques developed to perform inference on
probabilistic programs to calculate PS scores.  We build gray-box models of the
machine learning techniques by profiling actual training runs of the library,
and using profiled values to build abstract models of the training process.  We
amortize work across PS computations of different training points.
Probabilistic programs allow us to carry out these optimizations and reason
about them as program transformations.
\item We have built a tool \thetech implementing the approach for logistic regression and
boosted decision trees. \thetech is able to identify root causes of
misclassifications in several interesting data sets. We hypothesize that this
approach can be generalized to other machine learning tasks as well.
\end{itemize}

\section{Overview}
\label{Section:Overview}
We motivate our approach through the experience of Alice, a typical developer
who uses machine learning.

\begin{figure}[t]
{\begin{center}
\includegraphics[scale=0.4]{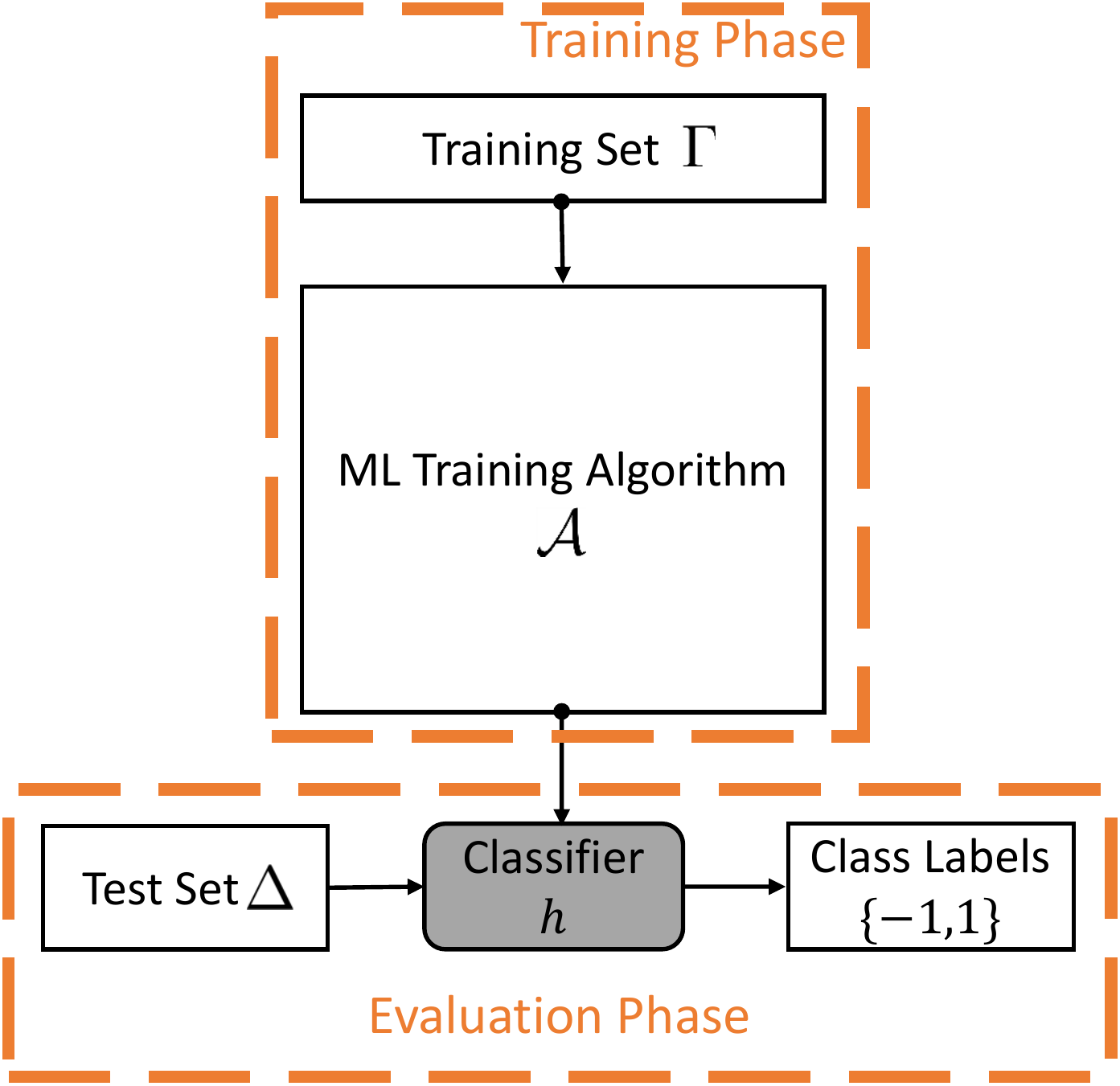}
\end{center}}
\caption{A two-stage design flow of a machine learning task: \emph{training phase} in which the 
ML algorithm $\cal A$ is applied to training set $\Gamma$ to learn classifier $h$, and 
\emph{evaluation phase} to judge the quality of $h$ on test set $\Delta$.}
\label{fig:ml-flow}
\end{figure}

\subsection{Typical Scenario}
\label{sec:typical-scenario}
Alice is not a machine learning expert, but needs to write a classifier for
images of vehicles and animals.  Mallory is a machine learning expert who
built a classification library using state of the art machine learning
techniques.  Alice decides to use Mallory's library, and since machine learning
libraries are driven by data, she carefully collects some amount of training
data $\{x^i\}_i$ with images of cats, dogs, elephants trucks, cars, buses etc.,
with labels $y^i = -1$ or $y^i = 1$, stating whether an image is that of a
vehicle or an animal respectively. She partitions it into a \emph{training set}
$\Gamma = \{(x^i,y^i)\}_{i=1}^M$, and a \emph{test set} $\Delta$, and picks out her
favorite ML algorithm, logistic regression, to learn a binary classifier that
separates vehicles from animals.

Alice runs Mallory's impeccable implementation of logistic regression on her
training set $\Gamma$ to learn the classifier $h : \Gamma \rightarrow \{-1,1\}$.  She then
evaluates $h$ over the test set $\Delta$.  This two-stage design flow is common, and
is shown in Figure~\ref{fig:ml-flow}.
Alice is happy with most classifications
being correct, but unhappy that a particular image $x_\Delta$ of a small car has
been incorrectly classified as an animal. She wants to find out an explanation
for why $h(x_\Delta) \neq y^t$.  She suspects some training data may be mislabeled,
so she wants to know what set of training instances $\Gamma_{x_\Delta} \subseteq \Gamma$
``caused" $x_\Delta$ to be misclassified.

\begin{figure}[t]
{\begin{center}
\includegraphics[scale=0.3]{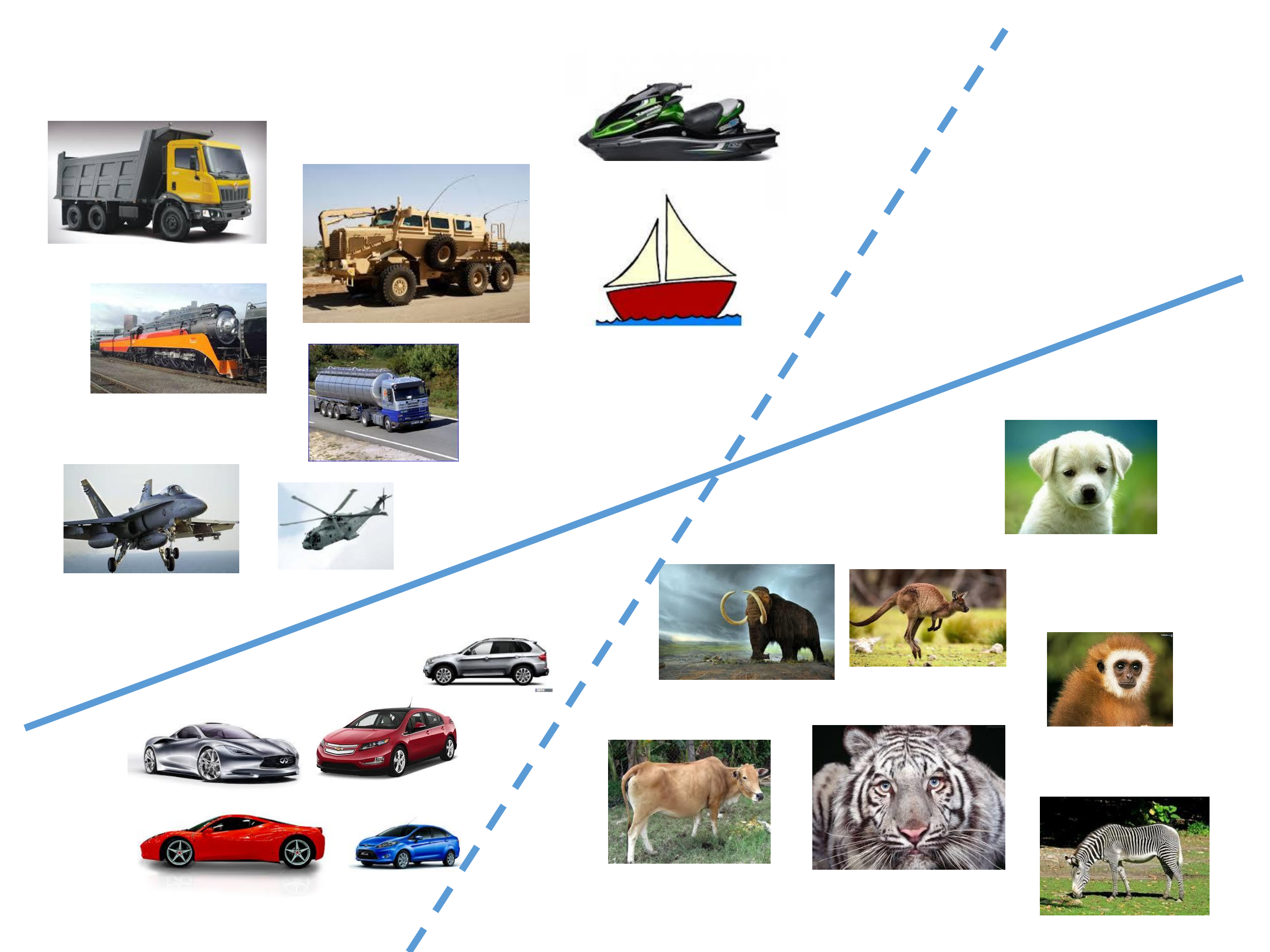}
\end{center}}
\caption{A classification example. All cars are incorrectly labeled as animals in the training set.
Thus the learned classifier (shown as a solid line) incorrectly classifies cars in the test data set as animals.
If the errors in the training set are fixed, then a correct classifier (shown in dotted line) will be learnt.
Our goal is to identify these errors in the training set efficiently.}
\label{fig:classifier}
\end{figure}

The cause for misclassification in her case is that each of the 5 cars
in her training data are classified as animals due to an error in her script
which collected the training data, as shown in Figure~\ref{fig:classifier}.
Since her test instance is a car, the classifier incorrectly classifies
it as an animal.

Alice's tried-and-tested approach to debugging is to start at the point of the error and work
backward and find which portions of the code led to the error~\cite{weiser}. 
Logistic regression uses the training data to
find a model $\theta$ which is an $n$-dimensional vector, where $n$ is the
number of features in the training data.  Once the model $\theta$ is calculated
(using training data), on the  test input $x_\Delta$, the output is given by $h(x_\Delta) =
\frac{1}{1+e^{-\theta^T\cdot x_\Delta}}$, which is a sigmoid function applied to the
dot product $\theta^T\cdot x_\Delta$ (see Sections~\ref{sec:lr} for more details). Thus, the
output for the test point $x_\Delta$ depends on {\em all} the indices of the $\theta$
vector, and it turns out that $\theta$ depends on {\em all} the elements of the
training set.  Thus, unfortunately, tracing backward through program
dependences does not enable Alice to narrow down the root cause of the failure.

Next, Alice turns to experimental approaches to answering why the 
classifier $h(\cdot)$ classifies the car $x_\Delta$  as an animal ~\cite{Zellerbook}. 
She picks training data points at random, and changes their labels and looks at 
the outcome after rerunning logistic regression. 
She finds that changing the label of a single training point makes no difference
to the classifier. Hence, she chooses subsets of training points, changes all
their labels simultaneously, reruns the training phase, and observes if the resulting (changed) classifier
correctly classifies the car $x_\Delta$.
She finds that switching the label on sets of training data points that include several cars
has much greater influence on the classification of $x_\Delta$ than other sets of points. 
Given that combinations of causes are involved, she wonders if there is a
systematic way to identify all (or most of) the possible causes.  Each
black-box experiment Alice runs is time consuming, since training is expensive.
Thus, Alice wonders if there is a better way to perform her experiments more 
systematically and more efficiently.

\subsection{Our Approach}
Our goal is to identify root causes by systematically and efficiently carrying
out the experimentation approach that Alice attempts to do.  There are two main
difficulties that Alice faces in her experimentation:
(1) Changing any single training instance does not fix the error. On the other
hand, trying to identify subsets of training points which fix the error is
infeasible due to the explosively large number of possible subsets of points.
(2) Each experiment takes a long
time to run, since it reruns the training algorithm from scratch.  We consider
each of these difficulties in detail.

In order to counter the first difficulty, instead of considering sets of
training points, we measure influence of individual points in causing the
classification. We use Pearl's theory of probabilities of
causation~\cite{ProbCausation} which formalizes the influence which a training
instance $t$ has on $x_\Delta$ being misclassified by considering all possible
labellings of the training points as possible worlds, and measuring the number of
worlds in which changing the label of $t$ changes the outcome of classification
of $x$.  In this spirit, we perform experiments with a large number of
alternate worlds, and for each world $w$ we can evaluate if in world $w$,
changing the label of $t$ influences the classification of $x_\Delta$.  Given a world $w$, we perform
this experiment simultaneously for all training points, and record which of the
training points influence the classification of $x_\Delta$ in world $w$.  We repeat the entire
experiment with several alternate worlds, and compute an aggregate score for each
training point $t$ on $x$ over all these alternate worlds. Such an aggregate
score is called Probability of Sufficiency or PS (see 
Section~\ref{sec:probabilistic-counterfactual-causality} for more details).

The second difficulty is the cost of building a model for each experiment (also
known as training).  Black-box experimentation does not scale because there are
a huge number of training points, and running the entire implementation of the
machine learning algorithm for each experimental trial takes a lot of time.  
We can potentially take a white-box view since we have access to its
source; however, incorporating the full details of the source poses a
scalability issue to even the state-of-art inference techniques.  
As a practical compromise, we take a gray-box view of the machine learning implementation,
and profile key intermediate values (that are hand-picked for each machine learning
algorithm) during the initial training phase.
Using the profiling information, we are able to use
intermediate values from runs of one training set  $\Gamma$ to efficiently calculate
the PS value for a related training set $\Gamma'$, where $\Gamma$ and $\Gamma'$ differ in a
small number of labels.  The exact values profiled, and the nature of the
gray-box model depends on the specific machine learning algorithm. In Section~\ref{sec:Debugging},
we show gray-box models for both logistic regression and decision trees.

\paragraph{Probabilistic Program for PS computation.}
We model the computation of the PS score as a {\em probabilistic
program}~\cite{fose}.  
Writing down the PS computation as a
probabilistic program enables us to use techniques developed for
probabilistic program inference to calculate PS.  It also enables to
concretely represent and reason about various optimizations to calculate PS as
program transformations. The probabilistic program we write for the PS score (see Section~\ref{sec:probabilistic-programs})  
models a Bernoulli distribution with mean given by the PS value, and inference over the probabilistic program
gives us an estimate of the PS value.

\paragraph{Empirical Results.} We have implemented our approach
in the tool \thetech.  In Section~\ref{sec:Evaluation}, we show empirical evaluation of \thetech using
synthetic benchmarks as well as real-world data sets.
We introduce 10\% random errors in the training data for synthetic benchmarks, and
systematic 10\% errors in the training data for real-world data sets.
We find that \thetech is able to identify a significant fraction of the systematically
introduced errors using the calculated PS scores. For randomly introduced errors,
the performance of \thetech depends on the nature of the benchmark. For instance, if the benchmark
already has a lot of inherent noise, then adding extra noise does not really change 
the classifier, and hence there is not enough information to perform root causing.
We also evaluate the effect of changing the labels for training points with high PS scores 
on the validation score. We find that the validation score improves monotonically
as we make changes for points with high PS scores, and starts to degrade if we change
labels for PS scores below 0.28, confirming that higher PS scores are more likely
candidates for root causing errors. We also find that using PS values from
multiple test points enables \thetech to find more root causes, as well as
improve validation score. See Section~\ref{sec:Evaluation} for more details.

\cut{
\paragraph{Probabilistic Model.}
The machine learning task above can be modeled as the following simple probabilistic program:
\begin{lstlisting}
T := inputTrainingSet;
D := inputTestSet;
Alg := LogisticRegression;

//training phase
h := train(Alg,T); //train classifier over T

//evaluation phase
D_bar := {}; //misclass. instances of D
for (X_d,Y_d) in D:
    if (h(X_d) != Y_d): //misclassification?
        D_bar := (X_d,Y_d)::D_bar;
score := 1-size(D_bar)/size(D);
return score;//return precision_score
\end{lstlisting}
The problem of root cause analysis is then given an instance
$d:(\vec{X}_d,Y_d)$ of \texttt{D\_bar}, could you find a subset $T_d \subseteq
\texttt{T}$ which is the cause for $h(\vec{X}_d) \neq Y_d$.  Thus, the problem
can be viewed as a (probabilistic) inference problem of identifying the set
$T_d$ based on the \emph{observation} $h(\vec{X}_d) \neq Y_d$, where the
probabilistic nature comes in through the input datasets $T$, $D$ and the
stochastic algorithm \texttt{Alg}.

The complexity of the model (and the problem), therefore, lies entirely within
the \texttt{train(Alg,T)} function and the level of detail of the training
algorithm \texttt{Alg} modeled by this call.
}

\section{Preliminaries}
\label{Section:Preliminaries}
In this section, we lay out the formal details of the machine learning tasks we
consider.  We introduce the intuition and formalization of the probabilistic
causality framework we propose.  We also introduce probabilistic programs, our
specification mechanism for probabilistic causality.

\subsection{Machine Learning Applications}

We consider classification tasks in supervised machine learning~\cite{bishop}.
Formally, a \textit{machine learning task} $(\Gamma, \Delta,{\cal A})$ consists of a training set 
$\Gamma = \{(x^i, y^i)\}_{i=1}^M$, and a test set $\Delta = \{(x^j, y^j)\}_{j=1}^N$
of labeled samples $(x^j, y^j)$, where $x^i, x^j \in
\mathbb{R}^n$ are called \emph{feature vectors}, $y^i, y^j \in \{-1,1\}$ are
called classification \emph{labels}, and $\cal A$ is an ML classification algorithm. 
Generally, developing a machine learning task consists of two phases as shown in
Figure~\ref{fig:ml-flow}:
\begin{enumerate}
\item \emph{Training Phase}, in which the training algorithm $\cal A$ is applied
  to the training set $\Gamma$ to derive as output a binary classification
  function $h : \mathbb{R}^n \rightarrow \{-1, 1\}$ that maps each 
  feature vector $x^i \in \mathbb{R}^n$ to a classification label $-1$ or $1$.
\item \emph{Evaluation Phase}, in which $h$ is applied to each sample $d$ in
  $\Delta$. The result of this phase is an evaluation score $S$ that is defined as follows:
\[S \equiv \sum_{(x^d, y^d) \in D}I(h(x^d)\neq y^d),\]
where $I(\varphi)$ is the $0$-$1$-indicator function.
\end{enumerate}
The goal of a machine learning task is to compute a classification
function $h$ over $\Gamma$ that generalizes well to the unseen test set $\Delta$ so as to
minimize the evaluation score $S$.  

\subsection{Probabilistic Counterfactual Causality}\label{sec:probabilistic-counterfactual-causality}
Before we formalize the definition of causality, we motivate it with an example.

\begin{example}
Consider a scenario where people vote for a government; choices are $A$ or
$B$.  In the final count of votes, out of $101$ people who voted,  $56$ voted
for $A$ and $45$ for $B$.
Everyone who voted for $A$ contributed to $A$'s winning; however, no individual
voter $v$ appears to be the sole cause (affects the outcome of the election).
The theory of causality via counterfactuals~\cite{CausalityBook,ProbCausation} proposes a solution to this
apparent predicament by considering alternate worlds in which an individual
$v$'s vote indeed affects the outcome. Informally, $v$'s vote is a cause if
there exists an alternate world (in this case an assignment to other voters'
choices) such that in this alternate world, the outcome is decided by $v$'s
vote alone. Such alternate worlds are called
\emph{counterfactuals} (counter to the actual world we observed).
 \cut{In other words, had $v$ voted for $B$ then $B$ would have won and $A$ would have won otherwise.} 

While considering the existence of counterfactual worlds helps establish
causality in a qualitative sense, we find it useful to consider a quantitative
measure of causation.  For example, consider the US presidential election where
different states have different numbers of electoral votes. California has
$55$ electoral votes, whereas Wyoming has only $3$ electoral votes. 
Intuitively, California voting for $A$ is a \emph{more significant} cause for $A$ winning than 
Wyoming voting for $A$. This notion can be captured by considering the number of
counterfactual worlds in which the outcome is affected. The number of alternate
worlds in which California affects the outcome of the election (any
world in which the difference in votes, barring California's, is less
than $55$) is greater than the number of alternate worlds in which
Wyoming affects the outcome of the election. We next focus on formally
stating this definition.
\label{Example:voting}
\end{example}

\subsubsection{Causal Models}

Informally, a \emph{causal model}\footnote{Our presentation is a
  simplification of the structural model semantics of causality that is
  suitable for our application. We refer the reader to \cite{CausalityBook} for a
  comprehensive introduction to counterfactual causality.} is a relationship
  between the inputs to any system and its output. A simple model for the
  voting example with $n$ voters is to have input variables $Y_1, \ldots , Y_n$
  (one for each voter), where each $Y_i \in \{-1, 1\}$, denoting a
  vote for $A$ or $B$ respectively.  Assume each vote $Y_i$ carries weight
  $a_i$, the causal model relating the inputs to the output (the predicate
  ``Has $A$ won?'') is therefore: \[a_1Y_1 + \cdots + a_nY_n > 0\]

\begin{definition}
A \emph{causal model} $M$ is a set of random \emph{input variables} $\{Y_1,
\ldots, Y_n\}$, each taking values in domain $D$,
an output variable $X$ over some output domain $D'$, and a \emph{structural
equation} $X = f(Y_1, \cdots, Y_n)$ relating the output variable $X$ to the
input variables, through some well-defined (deterministic) function $f : D^n
\rightarrow D'$ .  An assignment $w$ to all input variables is called a
\emph{world}.\label{Definition:causal-model}
\end{definition}

To consider counterfactual possibilities, we need to express quantities of the
following form: ``The value $X$ would have obtained, in the world $w$, had $Y$
been $y$''. Our tool for expressing counterfactuals is an intervention. An
\emph{intervention} $\inter{Y_i}{y_i}$ is a substitution of $y_i$ for the value
of $Y_i$ in a world $w$.  Formally, we represent
$f(w\inter{Y}{y})$ using the notation $\cf{X}{w}{\inter{Y}{y}}$.

To reason about uncertainty about the inputs to the causal model,
we augment causal models with a probability measure over worlds.

\begin{definition}   A \emph{probabilistic causal model} $(M, p)$ is a causal model $M$
  augmented with a probability measure $p$ over the input variables, i.e., a function
  from $D^n \rightarrow [0,1]$, where $D$ is the domain of each of the $n$ input variables.
\end{definition}

\noindent
As the structural equations are deterministic, $p$ also defines a distribution over the output $X$
of the causal model $M$, where:
\[Pr(X = x) = \sum_{\{w \mid f(w) = x\}} p(w)\]

\noindent
Similarly, we can define the probabilities of counterfactual statements:

\[Pr(\cfR{X}{\inter{Y}{y}} = x) = \sum_{\{w | \cf{X}{w}{\inter{Y}{y}} = x\}} p(w)\]

\noindent
We can also define the probabilities of conditional counterfactual statements as follows:

\begin{equation}
  \begin{array}{l l}
&    Pr(\cfR{X}{\inter{Y}{y}} = x \mid X \not= x, Y \not= y)\\ \\
=& \dfrac{Pr(\cfR{X}{\inter{Y}{y}} = x, X \not= x, Y \not= y)}{Pr(X \not= x, Y \not= y)}\\ \\
=& \displaystyle\sum_{\{w \mid \cf{X}{w}{\inter{Y}{y}} = x\}} p(w \mid X\not=x, Y\not=y)\\
\end{array}
\label{eq:cond-counterfactual}
\end{equation}

This is the conditional probability  of ``$X$ being $x$ when $Y$ is $y$, given
$X$ is not $x$ and $Y$ is not $y$''. This conditional probability may appear to
be zero at first glance, but under the counterfactual interpretation, $X$ and
$\cfR{X}{\inter{Y}{y}}$ are actually evaluated under different assignment for the
input variable $Y$. In fact, this conditional probability, known as
\emph{probability of sufficiency}, is the quantitative measure we use to
measure causality in our setting.

\paragraph{Probability of Sufficiency.} Assume that in the world $w: Y_1 = y_1, \ldots , Y_n = y_n$, 
it is the case that $X = x$.  Then the probability of sufficiency (PS) of some $Y_i = y_i$ 
being the cause for
$X = x$ is defined as:

\begin{equation}
PS(Y_i) \equiv Pr(\cfR{X}{\inter{Y_i}{y_i}} = x \mid X \not= x, Y_i \not= y_i)
\label{eq:ps}
\end{equation}

According equation \eqref{eq:cond-counterfactual}, the probability of
sufficiency measures the probability of each world in which $X \not= x$ and
$Y_i \not= y$, but changing $Y_i$ to its true value $y_i$ results in $X$
changing to $x$.  For the voting example, where each $Y_i$ represents an
individual vote, and $X$ represents the outcome of the election, these worlds
are ones in which an individual vote affects the outcome of the election.
Therefore, the probability of sufficiency for a voter $n$ who voted for $A$, is
the probability that $a_1Y_1 + \cdots + a_{n-1}Y_{n-1} + a_n > 0$ given $a_1Y_1
+ \cdots + a_{n-1}Y_{n-1} - a_n < 0$. Intuitively, for a larger number of
electoral votes  $a_n$, the set of worlds that satisfy the above condition is
greater, thereby leading to a greater $PS$ value for California with
$55$ electoral votes than Wyoming with $3$ electoral votes.

\subsection{Probabilistic Programs}\label{sec:probabilistic-programs}
\textit{Probabilistic programs}~\cite{fose} are ``usual'' programs with two additional 
constructs: (1) a \textit{sample} statement, that provides the ability to draw values from distributions, 
and (2) an \textit{observe} statement that provides the ability to condition on the values of variables.
The purpose of a probabilistic program is to implicitly specify a probability distribution.  
{\em Probabilistic inference} is the problem of computing an explicit
representation of the probability distribution implicitly specified by
a probabilistic program. 

\begin{figure}[t]
\begin{center}
\begin{tabular}{c}
\begin{lstlisting}
$\mathit{TwoCoins}()$
1:  $c_1$ := $\mathit{sample}(\mathit{Bernoulli}(0.5))$;
2:  $c_2$ := $\mathit{sample}(\mathit{Bernoulli}(0.5))$;
3:  $\mathit{observe}(c_1 \lor c_2)$;
4:  $\mathit{return}(c_1, c_2)$;
\end{lstlisting}
\end{tabular}
\end{center}
\caption{A simple probabilistic program.}
\label{fig:simplepp}
\end{figure}

Consider the probabilistic program shown in Figure~\ref{fig:simplepp}. This program tosses
two fair coins (simulated by draws from a Bernoulli distribution with mean $0.5$ in lines 1
and 2), and assigns the outcomes of these coin tosses to the Boolean variables $c_1$ and $c_2$. 
The $\mathit{observe}$ statement in line 3 blocks all executions of the program that do
not satisfy the condition specified by it (the Boolean expression $(c_1 \lor c_2)$). The meaning
of a probabilistic program is the probability distribution over the expression returned by 
it. For our example program $\mathit{TwoCoins}$, the distribution specified is the distribution over the 
pair $(c_1, c_2)$:  $Pr(c_1=0, c_2=1)$ $=$ $Pr(c_1=1, c_2=0)$ $=$ $Pr(c_1=1, c_2=1) = 1/3$, and
$Pr(c_1=0, c_2=0) = 0$.

\cut{

In words, in a world where $Y_i$ was set to some other value $y'_i \neq y_i$,
$X$ does not take on the value $x$, $PS$ is the probability of $X$ becoming $x$
when $Y_i$ is changed back to $y_i$ (through an \emph{intervention}).

Notice this definition does \emph{not} express a change in the degree of belief
about any \emph{one} specific world, but instead considers \emph{multiple}
different worlds.
The notion of a world here is deeper than that of an \emph{environment}
or a \emph{context} as it carries with it structural information similar to
control flow.}

\section{A Probabilistic Program to Compute PS}\label{sec:prob-prog-ps} 

In this section, we show how to encode the computation of the PS value as a probabilistic program.
Formally, we are given:
\begin{enumerate}
  \item A machine learning task $(\Gamma, \Delta, {\cal A})$, with training set
  $\Gamma = \{ (x^i, y^i) \}_{i=1}^N$, test set $\Delta = \{(x^j, y^j)\}_{j=1}^M$ and machine 
  learning algorithm $\cal A$.
  \item An error $\varphi(c)$, which is a predicate on
    the classifier $c$ produced by $\cal A$ (when run on training data $\Delta$).
\end{enumerate}

\begin{figure}[t]
\begin{center}
\begin{tabular}{c}
\begin{lstlisting}
$PS_i({\cal A}, \Gamma, \varphi)$ 
1:  $\Gamma.Y$ := $\mathit{sample}(p_T)$;
2:  $\mathit{observe}(\Gamma.Y_i \neq y^i)$;
3:  $c$ := ${\cal A}(\Gamma)$;
4:  $\mathit{observe}(\neg\varphi(c))$;
5:  $\Gamma.Y_i$ := $y^i$;
6:  $c'$ := ${\cal A}(\Gamma)$;
7:  $\mathit{return}(\varphi(c'))$;
\end{lstlisting}
\end{tabular}
\end{center}
\caption{A probabilistic program that specifies $PS(Y_i)$.}
\label{fig:psi}
\end{figure}

\noindent
An example of a predicate $\varphi(c)$ is $c(x^t) \not= y^t$, where $(x^t, y^t) \in \Delta$.
Another example predicate is that the total number of misclassifications over the entire test set $\Delta$
is less than some threshold value.
For each training instance $(x^i, y^i) \in \Gamma$, we wish
to measure the probability of sufficiency of the label $y^i$ for causing the error $
\varphi(c)$.

We first define a causal model for $\varphi(c)$. With each
training label $y^i$, we associate an input random variable $Y_i$ of the model.
Thus, lower-case variables will $y^i$ denote instances (or samples), and 
upper case variables $Y_i$ the corresponding random variable.
The probability distribution $p_{T}(Y_1, \ldots, Y_N)$  represents prior
beliefs over the training labels, and is based on the training set $\Gamma$.
The structural equation $f$ is simply the machine learning algorithm $\cal A$.

A succinct way of representing $PS(Y_i)$ is given by the probabilistic program shown in
Figure~\ref{fig:psi}.  The notation $\Gamma.Y$ represents the labels in the training set $\Gamma$.
Line 1 reassigns the labels in $\Gamma$ to a new sample of labels from the distribution $p_T$.  
Each assignment to $\Gamma.Y$ represents a different world (see Definition~\ref{Definition:causal-model}).
The observe statements in lines 2 and 4 reflect the counterfactual essence of the probability 
of sufficiency. They correspond to the conditions $Y\not=y$ and $X\not=x$ in Equation~(\ref{eq:cond-counterfactual}).
Line 5 models the intervention where $Y_i$ is set to $y_i$, and line 6 reruns the learning algorithm
$\cal A$ with the changed training set (where $Y_i$ has been updated).
The return value of this probabilistic program is a Bernoulli random variable
that expresses precisely the quantity $PS(Y_i)$ as defined in Equation (2).  
Specifying $PS(Y_i)$ as a probabilistic
program allows us to leverage recent advances in inference techniques for
probabilistic programs, as well as apply various program transformations to enable efficient
and scalable inference.

\section{Implementing Computation of PS}
\label{sec:Debugging}

In Section 4, we showed that computation of PS can be encoded as a probabilistic
program. We can thus implement computing PS using recent advances in
inference techniques for probabilistic programs~\cite{stan,r2}.
These techniques estimate the posterior probability distribution of probabilistic programs
using sampling. However, directly performing inference for the probabilistic program in
Figure~\ref{fig:psi} is intractable for the following reasons:
\begin{enumerate}
  \item We potentially need to consider all subsets ($2^N$) of a large number $N$ of
training points. Even for modest ML tasks, $N$ is typically greater than $1000$.
  \item Moreover, Figure~\ref{fig:psi} requires inference to be performed once for each label $Y_i$, where the
number of training labels can be very large.
  \item For each subset, inference needs to be performed over the full implementation details of the training algorithm $\cal A$.
Handling such highly optimized C++ code (usually several hundred lines) is beyond the scalability of any existing sampling
 technique for probabilistic inference. Finally, $\cal A$ needs to be re-run for each relabeling.
\end{enumerate}
We tackle these three challenges by employing the following assumptions and techniques:
\begin{itemize}
\item \emph{Low noise level} - We assume that only a small number of training points are
    mislabeled. This is encoded in the distribution $p_T$ (in line 1 in Figure~\ref{fig:psi}) which decays fast as
    the number of mislabeled points grows. We can thus restrict our
    samples to a small portion of the space of all possible mislabeled subsets. This is implemented in the probabilistic
    program by choosing a prior distribution $p_T$ that is heavily biased toward the given training set $\Gamma$.
\item \emph{Sample reuse} - To address challenge 2,
we observe that samples drawn for one label $y_i$ can be
reused for other labels. We can think about this as the following transformation on probabilistic programs:
(1) move the observe statement from line 2 in Figure~\ref{fig:psi} after the observe statement in line 4; and (2) rewrite that
 observe statement as a conditional (shown in Figure~\ref{fig:parallel-psi}).
\begin{figure}[t]
\begin{center}
\begin{tabular}{c}
\begin{lstlisting}
$PS_i({\cal A}, \Gamma, \varphi)$
1:  $\Gamma.Y$ := $\mathit{sample}(p_T)$;
2:  $c$ := ${\cal A}(\Gamma)$;
3:  $\mathit{observe}(\neg\varphi(c))$;
4:  $\mathbf{if} (\Gamma.Y_i \neq y^i)$
5:    $\Gamma.Y_i$ := $y^i$;
6:    $c'$ := ${\cal A}(\Gamma)$;
7:    $\mathit{return}(\varphi(c'))$;
8:  $\mathit{return}(0)$;
\end{lstlisting}
\end{tabular}
\end{center}
\caption{An efficient probabilistic program that specifies $PS(Y_i)$.}
\label{fig:parallel-psi}
\end{figure}
Even though the programs in Figure~\ref{fig:psi} and Figure~\ref{fig:parallel-psi} are equivalent,
the program in Figure~\ref{fig:parallel-psi} performs an important optimization.
Since the lines 1--3 in Figure~\ref{fig:parallel-psi} are the same for all $i$, the work for executing
these statements can be shared across all $i$. Then, for each sample generated, after a quick check that condition
in line 4 is satisfied, we execute (in parallel for all such $i$) lines 5--7.

\item \emph{Model approximation \& Robustness} - Instead of operating over the complex implementation
of training algorithm $\cal A$, we approximate it suitably via an approximate causal model $\widehat{\cal A}$.
The key observation is that some ML algorithms are generally robust~\cite{elisseeff2001algorithmic}.
When a small subset of the input changes, the internal computations leading to the final
classifier do not change significantly. In fact, being insensitive to
small changes in inputs is desirable for machine learning algorithms to reduce overfitting.
Thus, we design approximate causal models for two widely used classification algorithms
logistic regression (see Section~\ref{sec:lr}) and decision trees (see Section~\ref{sec:dt}), which enable
efficient computation of $PS(Y_i)$ below.
\end{itemize}
Next, we present the essential details of our approximate causal models for Logistic Regression (Section~\ref{sec:lr})
and Boosted Decision Trees (Section~\ref{sec:dt}). \cut{In Section~\ref{sec:prob-prog-ps} we equated the causal model with the structural equations induced by $\cal{A}$.
As our approximate causal model $\widehat{\cal{A}}$ we take the constraints $\varphi(c)$ induced by $\cal{A}$ as a function of the $Y_i$ }

In Figure~\ref{fig:parallel-psi}, we need to run the entire algorithm $\mathcal{A}$
each time constraint $\varphi(c)$ is to be evaluated for a sample. We now describe
our approximate models $\widehat{\mathcal{A}}$ which are simpler equations for computing $\varphi(c)$ in terms of the training labels $\{Y_i\}_i$.
The details on deriving the constraints and general details on these algorithms can be found in the supplementary materials.

\subsection{Approximate Logistic Regression Model}\label{sec:lr}

Logistic Regression (LR) is a popular statistical classification model that
uses a sigmoid scoring function:

\[h(Z) = \frac{1}{1+e^{-Z}}\]

where $Z = \theta^T x$ is the product of inferred classifier weights
$\theta$ and the feature vector $x$ of some instance. When logistic regression
is used for binary classification, instances with score $h(\theta^T x) <
0.5$ are classified $c(x)= -1$ and those with score $h(\theta^T x) \geq
0.5$ as $c(x) = 1$.

The classifier $\theta$ is learnt through an iterative gradient descent process
that iteratively finds a better classifier $\theta$ to fit the data.
For Stochastic Gradient Descent, the classifier is improved by iterating over
the following update equation:

\begin{align}
\theta^K = \theta^{K-1} + \alpha^{K-1} \sum_{i = 1}^N y^i x^i h(y^i~(\theta^{K-1})^Tx^i),
\label{Equation:theta-K}
\end{align}
Here, $\alpha$ is known as the step size, and $N$ is the number of training instances.
The above vector equation represents an update for each component of $\theta^K$

By making a robustness assumption that score on training points in the
penultimate iteration $K-1$ of gradient descent, i.e. $h((\theta^{K-1})^T x^i)$
does not change greatly, it turns out that the following equation provides an
approximation for the final classifier $\theta$ on a different labelling of
training labels $\{Y^i\}_i$ (details are found in A.1 of supplementary
materials).

\[ \theta = \theta^{K-1} + \alpha^{K-1} \sum_{i = 1}^N Y^i x^i
h(y^i~(\theta^{K-1})^Tx^i) \]

Using this, we can simplify the condition $\varphi(c)$ of the classifier $c$
incorrectly classifying a test point $x$ as follows:

\begin{align}
\varphi(c)\ :\ &h_{\theta}(x) \geq \frac{1}{2}  \Longleftrightarrow \theta^T x \geq 0,\text{ and}\label{Equation:PhiGeq0}\\
&h_{\theta}(x) < \frac{1}{2} \Longleftrightarrow \theta^T x < 0,\label{Equation:PhiLeq0}
\end{align}
where $\theta^T x$ is simply the product:
\[ (\theta^{K-1} + \alpha^{K-1} \sum_{i = 1}^N Y^i x^i h(y^i~(\theta^{K-1})^Tx^i))^T x \]

Notice now that due to this simplification, the condition $\varphi(c)$ is a linear
constraint on the input random variables $\{Y_i\}_i$. To compute the coefficients
of each $Y_i$, we need to profile $\theta^{K-1}$ and the step size $\alpha^{K-1}$.

\subsection{Approximate Decision Tree Model}\label{sec:dt} Decision trees are
tree-shaped statistical classification models in which each internal tree node
represents a decision split on a feature value, and associated with each leaf
node, is score $s$. The score $s(x)$ of a particular point $x$ is computed by
evaluating the decision at each internal node and taking the corresponding
branches until a leaf is reached. The leaf score is returned as the score
$s(x)$.  Each leaf can be viewed as a region $R$ in the feature space and a
tree is a partitioning $\{R_1, \cdots R_L\}$.  In Gradient Boosted Decision
Trees~\cite{Friedman00}, instead of a single tree, a number of trees called
an \emph{ensemble} are iteratively learnt through a gradient descent process.
Informally, after $n$ trees have been learnt, the $(n+1)^{\text{th}}$ tree is
learnt by fitting a new decision tree on the \emph{error residual} $\bar{y}_i$
of each training label $y^i$. The error residual can be thought of as the
difference in the aggregate score due to $n$ decision trees learnt so far, from
the true label. This iterative process stops after a fixed number of learning
rounds.

Under a robustness assumption for scores for training points and the learnt
regions for each of the decision trees, it turns out that for an alternate
assignment $\{Y_i\}_i$ to the training labels, the score $s(x)$ for a
particular test point $x$ can be written as follows (see  supplementary material for details):

\[
  s(x) = s_0(x) + \sum_{n = 1}^L \sum_{k = 1}^{K_n} \sum_{i = 1}^N \frac{(Y_i - y_i) \sigma I(x_i\in R_{nk})I(x\in R_{nk})}
                {\sum_{x_i\in R_{nk}}|\bar{y}_{ni}|(2\sigma - |\bar{y}_{ni}|)}
\]

\noindent
Here, $L$ is the number of decision trees learnt in the ensemble, $K_n$ is
the number of leaves in the $n^{\text{th}}$ tree and $N$ is the number of
training instances. $R_{nk}$ is the region represented by the $k^{\text{th}}$
leaf of the $n^{\text{th}}$ tree and $\bar{y}_{ni}$ is the error residual for
for the $i^{\text{th}}$ training label after $n$ iterations.

While this might appear to be a very complicated computation, notice that it is
only a linear computation on $\{Y_i\}_i$ and all the coefficients can be
precomputed from a single run of the algorithm.  As a result, the condition
$\varphi(c)$ of test point $x$ being misclassified can be written as $s(x) \geq
0$ (resp. $< 0$), which is a linear constraint on the training labels $Y_i$.

Another important implication of this equation is that the coefficient of label
$Y_i$ is a multiple of $I(x_i\in R_{nk})I(x\in R_{nk})$.  Therefore, the
coefficient of $Y_i$, is nonzero only when the corresponding training point
$x_i$ and the test point $x$ belong to the same leaf in some tree $n$. Hence,
only a small number of training labels $Y_i$ actually have any affect on the
classification of a training point $x$.

\subsection{The \thetech{} debugging tool.}
We have implemented the computation of PS value using the above mentioned
optimizations in the tool \thetech.
\thetech{} takes a machine learning task $(\Gamma, \Delta, {\cal A})$, a set of
misclassified test points (bugs) $B \subseteq \Delta$ as input, and produces
root causes in the form of training labels as output.

Currently \thetech supports logistic regression and decision trees, and uses
an approximate model $\widehat{\cal{A}}$ of these machine learning algorithms
as described above.
\thetech runs a single execution of the industrial strength implementation of these machine learning
techniques. We have hand instrumented the implementation of both algorithms to
profile for specific parameters that make $\cal{A}$ and $\widehat{\cal{A}}$ operate in lockstep.

For logistic regression, we record the iterates of $\theta$ and line search steps $\alpha$
in the stochastic gradient descent variant of search algorithm. Additionally, we memoize classification
scores $h(\theta^Tx)$ and reuse across different training labels $Y^i$.

For boosted decision trees, we record the regions corresponding each individual
tree in the ensemble and error residuals $\bar{y}_i$ for each iteration
of gradient descent.

The core engine driving \thetech{} is the probabilistic program together with
its sampling based inference engine (as defined in
Figure~\ref{fig:parallel-psi}). This program uses the label distribution
$p(\Gamma)$ over the training set $\Gamma$, the approximate model
$\widehat{\cal A}$ for the training algorithm $\cal A$ in order to compute the
PS scores for the training instances.

Training instances are sorted and highest ranked PS score instances
$T_{\varphi(c)} \subseteq T$ are returned as proposed causes. The user then
examines the instances in $T_{\varphi(c)}$ closely, fixes any labeling
inconsistency they detect, and re-runs $\cal{A}$ on the modified set $T$.
In Section~\ref{sec:Evaluation}, we show that flipping the labels for
training instances as-is without inspection provides improvements
in classifier performance.

t

\section{Evaluation}
\label{sec:Evaluation}

\cut{
Evaluation of the correctness of causal reasoning is nontrivial.  The
probabilistic program model we build directly encodes the definition of PS.
Thus, the sources of imprecision come in from the profiling approximation,
choice of prior distribution, and inference.

A practical evaluation strategy might be to introduce errors in the training
set and then run \thetech over new misclassifications introduced as a result of
training over this erroneous training set. One might expect to be able to
retrieve the added mistakes as the cause for misclassification. To see why this
is unreasonable to expect, let us return to the voting setting of
Example~\ref{Example:voting}.  Recall the scenario, where $A$ was leading with
56 votes and $B$ had 45 votes. Analogous to introducing errors in the training
set, we pick 10 voters from $A$'s camp and change their vote to $B$, thereby
leading to $B$ winning.  Notice that these 10 voters are now indistinguishable
from any of the voters who originally voted for $B$ and therefore cause $B$'s
victory as much as any other $B$ voter.
}

\begin{figure}[t]
\centering
\includegraphics[width=0.95\linewidth]{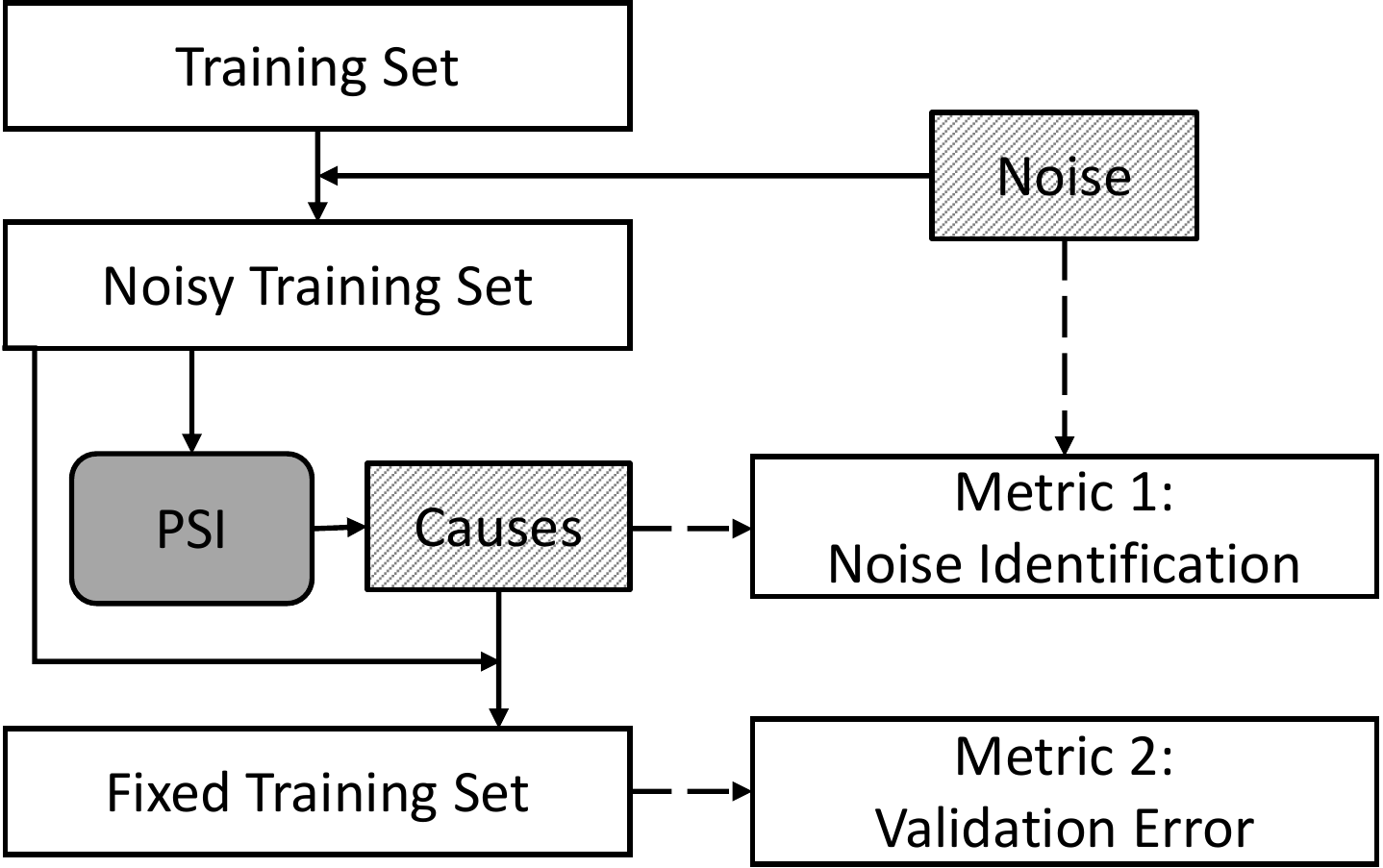}
\caption{Experimental setup for evaluation.}
\label{fig:workflow}
\end{figure}

In this section, we empirically evaluate the effectiveness of \thetech. All experiments
were performed on a system with a 1.8 GHz Intel Xeon processor and 64 GB RAM running Microsoft Windows 8.1.

We evaluate the applicability of \thetech with respect to two different
debugging requirements: (1) identifying errors in training data, and  (2)
reducing errors on unseen data. To evaluate these two metrics, we follow the
workflow described in Figure~\ref{fig:workflow}, where we first add noise to 10\%
of the training labels of a dataset. 
This perturbation introduces new misclassifications in the
test set.  We run \thetech on the new misclassifications with the goal of
finding the most likely training instances to cause the new misclassifications,
and make the following measurements:

\begin{enumerate}
	\item \textbf{Identifying Relevant Noise.} We measure the fraction of training
instances that \thetech returns that are known to have erroneous labels. 
	\item \textbf{Reducing Validation Error.} We introduce a separate \emph{validation set}
of instances independent of the training and test sets.  We measure the
accuracy of (i) a classifier learnt on the noisy training set, versus (ii) a
classifier learnt on the noisy training set with the most probable causes
suggested by \thetech flipped. We then measure the reduction in errors
in the second classifier with respect to the first.
\end{enumerate}

Table~\ref{tab:datasets} shows the datasets we study.
We consider two kinds of datasets:  (1) real-world data, and (2) synthetic data. 
The first two rows of the table are real-world datasets: \irl{sentiment}, an IMDB movie review
dataset \cite{CornellIMDB} used for sentiment analysis, and
\irl{income}, a census income dataset \cite{UCIDataset} used for predicting
income levels. 
The third and fourth rows are synthetic datasets:
\irl{2gauss} and \irl{concentric}, both produced by generating data from spherical Gaussian distributions.

We also consider two kinds of noise or errors in the datasets: (1) systematic noise, and (2) random noise.
We find that \thetech can identify systematic
noise with significant accuracy, thereby leading to a reduction
in classification errors on unseen data. We also observe that adding random
noise does not lead to a significant increase in errors, and therefore our
causal analysis does not have any room to identify errors. The results of these experiments are
summarized in Table~\ref{tab:experimental-results}, and explained in detail in
Sections~\ref{sec:eval:syst} (Systematic Noise), and \ref{sec:eval:rand} (Random Noise).
Additionally, we observe that combining information from multiple
misclassifications leads to better noise identification.

\begin{table}
\begin{center}
    {\footnotesize
   \begin{tabular}{|r||c|c|c|c|}
   \hline
   \\      Dataset          & Features & Training & Noise &  Source\\
\\   \hline
   \irl{sentiment}  & 13132 & 2000 & Syst. -- 10\% & Real\\
   \irl{income}     & 122   & 4000 & Syst. -- 10\% & Real\\
   \irl{2gauss}     & 2     & 1000 & Rand. -- 10\% & Synth.\\
   \irl{concentric} & 2     & 2000 & Rand. -- 10\% & Synth.\\
   \hline
   \end{tabular}}
\end{center}
\caption{Summary of the benchmark datasets showing the number of features, training instances, noise profile added, and 
source of each dataset.}
\label{tab:datasets}
\end{table}

\newcommand{\valerrs}[3]{#1 {\tiny{$\rightarrow$}} #2 {\tiny{$\rightarrow$}} #3}

\setlength{\tabcolsep}{3pt}
\hskip-0.5cm
\begin{table}
\begin{center}
    {\footnotesize
   \begin{tabular}{|r||c|c|c|c|c|}
   \hline
   \multirow{2}{*}{Dataset}  & \multicolumn{2}{c|}{Accuracy} & \multicolumn{2}{c|}{Validation Error} & \multirow{2}{*}{Time}\\
   \cline{2-5}
                                     & LR & DT & LR & DT   &  \\
   \hline
   \irl{sentiment}     & 0.58  & 1.00 & \valerrs{0.14}{0.24}{0.19}   & \valerrs{0.18}{0.26}{0.21} & 24m\\
   \irl{income}        & 0.70  & 1.00 & \valerrs{0.19}{0.25}{0.23}   & \valerrs{0.18}{0.24}{0.19} & 58m\\
   \irl{2gauss}        & 0.93  & 0.38 & \valerrs{0.01}{0.03}{0.02}   & \valerrs{0.01}{0.02}{0.02} & 12m\\
   \irl{concentric}    & 0.11  & 0.18 & \valerrs{0.47}{0.49}{0.48}   & \valerrs{0.09}{0.10}{0.10} & 14m\\
   \hline
   \end{tabular}}
\end{center}

\caption{Summary of experimental results. 
For each dataset, we report the accuracy (which is the fraction of errors introduced
that are identified by \thetech), 
, change in validation error for both logistic regression (LR) 
and decision trees (DT) and corresponding \thetech runtimes in minutes. Validation errors
are for a sequence of classifiers learnt on noise free dataset, noisy
dataset, noisy dataset with \thetech's suggestions flipped.}

\label{tab:experimental-results}
\end{table}

\subsection{Systematic Noise}
\label{sec:eval:syst}

We add systematic noise to the \irl{sentiment} and \irl{income} datasets, simulating systematic
errors in the data collection process.  For \irl{sentiment}, we pick a word that appears in
10\% percent of instances, and mark these as negative reviews. For
\irl{income}, we pick a demographic that covers 10\% of the population, and mark
them as high income.

When noise is added, the validation error increases. For instance, 
in the \irl{sentiment} dataset, with logistic regression, the validation error
increases from 0.14 to 0.24.
\thetech is able to successfully identify systematic noise in
datasets. For instance, in the \irl{sentiment} dataset, 
with logistic regression, \thetech
returns as causes, 58\% are points that were incorrectly labeled in the
dataset. We call this number accuracy. For decision trees 
the accuracy is 100\%!
The accuracy results for \irl{income} benchmark are even better---70\% with logistic regression and 100\% with decision trees.
Note that for datasets with 10\% noise, the baseline
for randomly picking noise would yield an accuracy of 10\%. We discuss the reasons
behind the difference in accuracy for the two algorithms in Section \ref{sec:eval:lrvdt}.

For systematic errors on a real world dataset, we find that using \thetech's
output to flip training labels leads to a reduction in validation error even
when not all the points suggested are true errors. For instance, for \irl{sentiment}
and the \irl{income} benchmarks, with decision trees,
validation error reduces by 0.05 and 0.05 respectively (which is a reduction of 20\%).

\subsection{Random Noise}
\label{sec:eval:rand}
In the two synthetic datasets \irl{2gauss} and \irl{concentric},
we randomly flip the labels for 10\% of the points as noise to the dataset.  We
use synthetic datasets to evaluate random noise, and rule out any preexisting
systematic noise.
We find that for random noise, for a very cleanly separable
dataset such as \irl{2gauss} (note that the initial validation error
for this benchmark is 0.01 with both classification schemes, a very low number), 
it is possible to identify noise with reasonable
accuracy (93\% for logistic regression, 34\% for decision trees). 

However, with an inherent noise such as in the \irl{concentric} dataset
(note that the initial validation error logistic regression is 0,47), added noise cannot
be distinguished from existing noise using causal measures. 
We observe that validation error changes very little when random noise is added to the 
dataset, and very few new misclassifications are introduced. This explains why our
causal analysis, which depends on observing
changes in outcome when inputs are perturbed, does not identify random noise 
correctly. However, in the presence of very low inherent noise, random noise 
``stands out'' and is easy to identify as is the case in \irl{2gauss}.

\subsection{Insights}
\begin{figure}[t]
\centering
\includegraphics[width=0.95\linewidth]{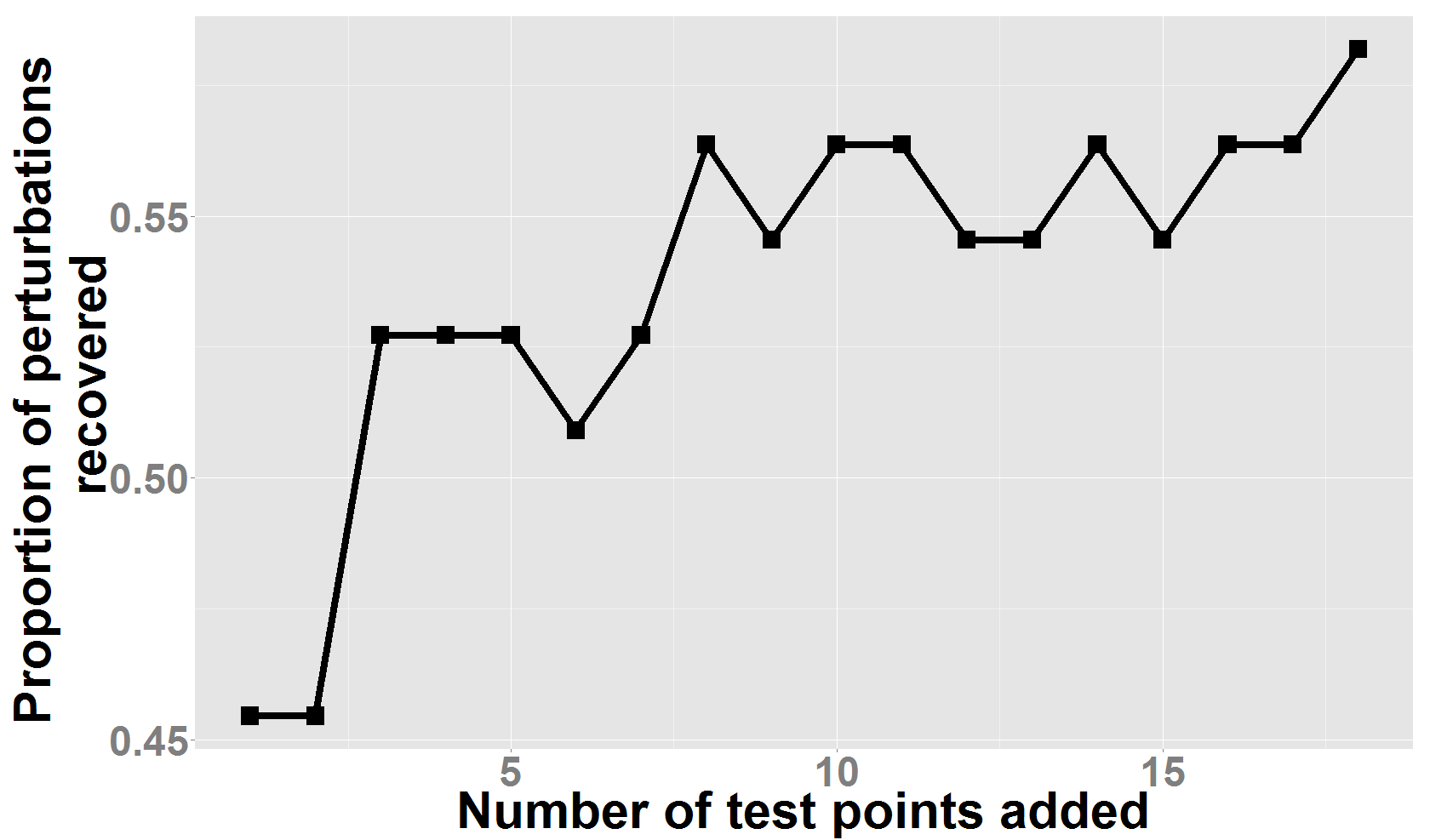}
\caption{Graph showing the improvement in noise identification for Logistic Regression runs on \irl{sentiment} dataset as more test points are added.}
\label{fig:noise}
\end{figure}

As shown in Figure~\ref{fig:noise}, we observe combining information from
multiple misclassified test points leads to better noise identification. We
combine information by choosing $\varphi(c) = \varphi_1(c) \land \cdots \land
\varphi_k(c)$, to be the predicate in our PS computation, where each
$\varphi_k(c)$ is an individual misclassification. For \irl{sentiment},
precision improves from 0.45 to 0.54 by adding 18 test points.
We find this encouraging---with more evidence of test errors, \thetech is able
to do better root causing, which indicates the robustness of the analysis.

\begin{figure}[t]
\centering
\includegraphics[width=0.95\linewidth]{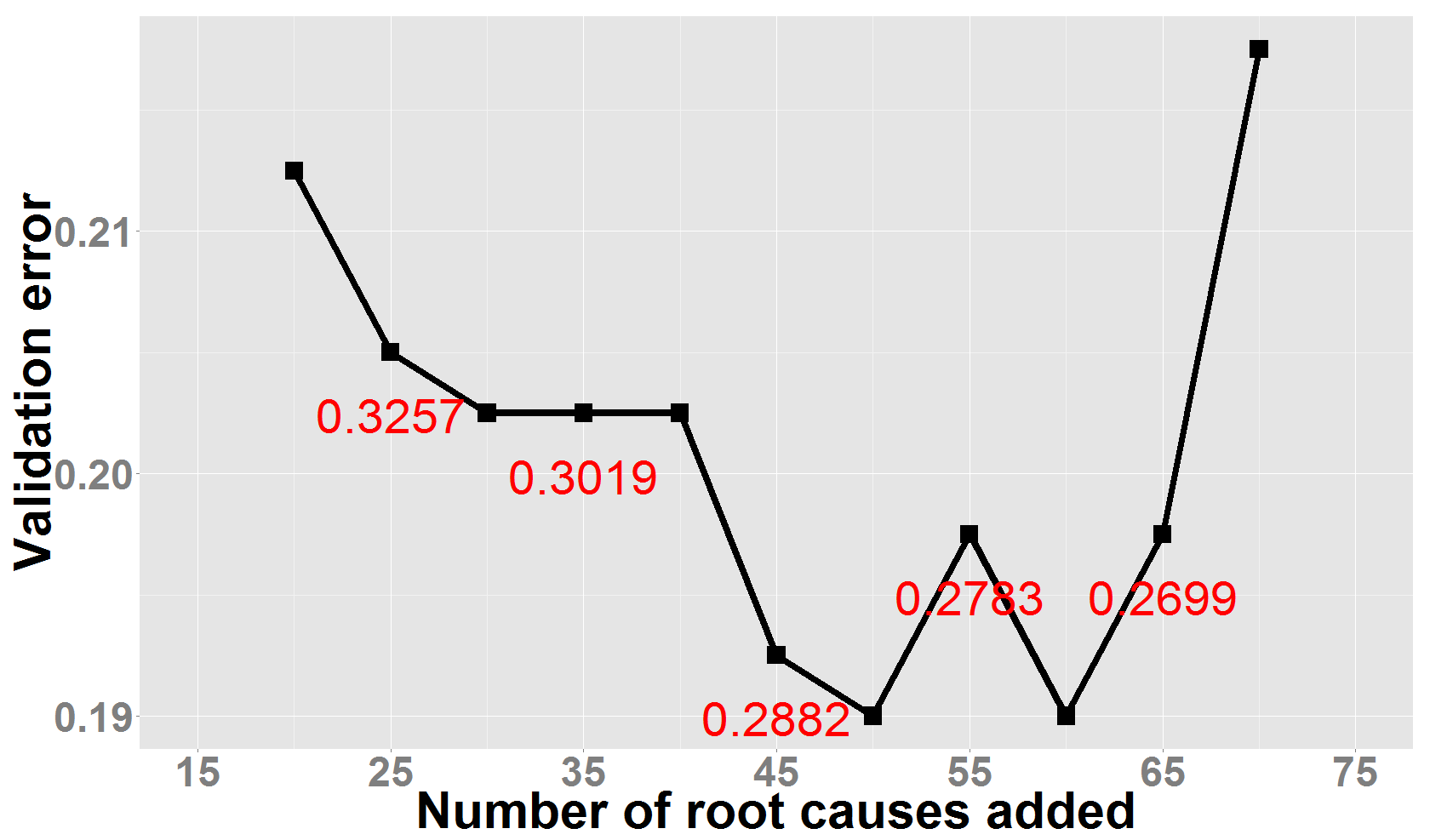}
\caption{Graph showing reduction in validation error for Logistic Regression as the number of root
causes added increases, together with PS threshold $\tau$ (in red) for that level.}
\label{Fig:validation-error}
\end{figure}

Next, we study how to interpret the PS score, and specifically what (absolute or relative)
value of the PS score indicates an error in the training point.
We study this by sorting training points based on the PS score, and measuring
the validation error as we flip more points starting with the point with the highest
PS score and iteratively flipping more points.
Figure~\ref{Fig:validation-error} shows an interesting trend: validation error
reaches a minimum at some PS threshold $\tau$, and changing labels below $\tau$
has a detrimental effect.  Plotting such a variation between validation error
and PS threshold as part of the debugging process could provide an empirical
method for choosing optimal PS thresholds.

\subsubsection{Logistic Regression vs. Decision Trees}
\label{sec:eval:lrvdt}

To explain why logistic regression and decision trees behave so differently
under our causal analysis, it is important to understand how training points
influence the score for a particular test point.  Logistic regression fits a
single hyperplane that best separates the training set. Therefore, changing the
label of any training point affects the hyperplane, and hence the score, but only 
to a small degree.
Decision trees, on the other hand, divide the feature space into smaller
regions, one for each tree in the ensemble. To compute a score for a test
point $\delta \in \Delta$, only training points that lie in the same region as $\delta$
contribute to the score of the test point. In effect, the actual set of points
that affect the outcome for a test point is much smaller for decision trees
than for logistic regression.

\section{Related Work}
\label{Section:RelatedWork}
We briefly review related work in the two most relevant areas: software debugging and counterfactual theory of causality.

\subsection{Software Debugging}

Debugging software failures is a very well-studied problem. In this section, we survey and compare existing
software debugging techniques with \thetech{}.

\paragraph{Program analysis.} Program slicing~\cite{weiser} computes a subset of statements in a program that can influence 
a variable at any program point. Since slicing can be used to effectively reduce the size of the code under analysis, it 
has many applications in software debugging~\cite{weiser-debugging}. Unfortunately, as described in Section~\ref{sec:typical-scenario}, 
slicing is not very effective for debugging machine learning tasks as the ``data slice" that influences any test error 
is usually the whole training set. In contrast, \thetech{} is based on a formal notion of causality introduced
by \cite{CausalityBook,ProbCausation}, and uses this effectively to isolate training data slices that are responsible
for test errors.

\cite{ball} use a software model checker to generate correct and incorrect traces, and compute the differences between
these traces to localize software defects. Unlike their work which looks for bugs in code, we look for bugs in training data.
However, they consider both correct and incorrect traces, while we consider only misclassified test data. There may be
a way to also make use of correctly classified test data, and we leave this idea for future work.

\paragraph{Delta-Debugging.} This debugging technique analyzes differences between failing and passing runs of 
a program to detect reasons for the failure~\cite{zeller2002isolating,zeller1999yesterday}. In this setting, a cause is defined to be the smallest part of 
a program state or the smallest part of the input, which when changed, converts a passing run into a failing run. 
As discussed in Section 2, the main challenge in our setting is that there are usually several causes (several
mislabels in the training data) for a misclassified test point, and it is not clear how to use delta-debugging
to search through all possible subsets of misclassified training labels. Instead, we use Pearl's PS score, which is very unique to our work.

\paragraph{Statistical debugging.} The CBI project and its variants~\cite{cbi1,cbi2,holmes} use information collected from 
program runs together with a statistical analysis in order to compute predicates that are highly correlated with failures. 
This technique also requires information from a large number of passing and failing runs of the program, and is applicable
when there are bugs in code, and a large number of users are using the same code, and triggering the same bug.
In contrast, we assume that the machine learning code is implemented correctly, and the bugs are in the training data.
In addition, each user has their own training data, which could have errors, which is a very different problem setting
than the one addressed by statistical debugging.

\subsection{Counterfactual Causality}

\thetech computes the \emph{probability of sufficiency} (PS) of a training
point in causing classification errors. PS, suggested by Pearl in
\cite{ProbCausation}, is part of a rich body of counterfactual theories of
causation first proposed by Lewis~\cite{Lewis1973}. In particular, in his
book~\cite{CausalityBook}, Pearl formulates structural equations as a
mathematical framework for reasoning about causality.  

\paragraph{Actual Causation.} An application of the structural equations framework is
Halpern and Pearl's definition of \emph{actual causation} \cite{HalpernP01}
(known as the HP-Definition). As in our setting, actual causation aims to
identify which input variables \emph{actually} caused some outcome. While the
HP-definition is able to explain a large variety of subtle issues around
causality, operationally, it is intractable to employ in our scenario, since to
verify whether an alternate world satisfies the definition, one needs consider
an exponentially large number of possibilities.  Additionally, the
HP-definition is qualitative, and does not suggest a quantitative measure of
causality, which is necessary for \thetech{} to rank most likely causes.

\cut{\paragraph{Counterfactual Causality.} Pearl and Halpern have developed a complex framework~\cite{},~\cite{},~\cite{} to formally determine causality in a counterfactual setting. To illustrate its applicability to reasoning about programs consider the following simple program:
\begin{lstlisting}[language=C]
        bool x, p, C, C1, D, prop;
        x := nondet();
        p := nondet();
        if (x):
            C1 := false;
            D  := true;
        else: 
            C1 := true;
            D  := false; 
        C := p && C1;
        if (C || D): 
            prop := false;
        assert(prop); 
\end{lstlisting}
The control flow of the program is deterministic except for the initialization of the two input variables \texttt{x}, \texttt{p} to nondeterministic boolean values.
Based on these values property \texttt{prop} ultimately holds or not.
  
Suppose we observe that the assertion in the program fails and we want to infer whether \texttt{x}, or \texttt{p} is the root cause for the failure. Weakest precondition reasoning yields the following result: $\texttt{x} \lor (\neg \texttt{x} \land \texttt{p})$, i.e. either \texttt{x} is a cause, or $\neg\texttt{x}$ and $\texttt{p}$ together cause the failure but we cannot answer the question one way or another. In the counterfactual causality framework, \texttt{x} is deemed the actual cause.
}

\section{Conclusion and Future Work}
\label{Section:Conclusion}
Unlike debuggers for coding errors that are ubiquitous, debuggers for data errors are less common.
We consider machine learning tasks, and in particular, classification tasks where incorrect classifiers can 
be inferred due to errors in training data.  We have proposed an approach based in Pearl's theory of 
causation, and specifically Pearl's PS (Probability of Sufficiency) score to rank training points that
are most likely causes for having arrived at an incorrect classifier.
While the PS score is easy to define, it is expensive to compute. Our tool \thetech employs several optimizations to scalably compute
PS scores including modeling the computation of the PS score as a probabilistic program, exploiting program transformations and efficient inference techniques for probabilistic programs, 
and building gray-box models of machine learning algorithms (to save the cost of retraining). Due to these optimizations, \thetech is able to correctly
root cause data errors in interesting data sets.

Our work opens up several opportunities for interesting future work.  An immediate
next step is to consider ML tasks involving algorithms such as support vector machines and neural
networks~\cite{bishop}. In order to do this, we need to design and develop scalable approximate models 
for these algorithms.

Another interesting direction is to support a wider class of causes.
Potential causes include identifying which feature in a test set causes a
misclassification, overfitting in the learning algorithm, incorrect parameters to the learning algorithm, and others.

A practical dimension to explore is scale. We have been able to study
datasets with thousands of points using \thetech. Industrial big-data systems
have millions of points, and more work needs to be done to scale root cause
analysis techniques to work at such scale.

\cut{
While our initial results are promising, this is only an initial step toward developing algorithms and 
tools for debugging data errors, and our work has several restrictions. 
We have considered only classification problems, and even within classification, 
we have considered only two types of classification algorithms---logistic regression and decision
trees. An immediate next step is to consider more algorithms such as SVMs (Support Vector Machines) and 
Neural Networks~\cite{bishop}. Our intuition is that our techniques should generalize to these 
algorithms, and we plan to explore these in future work. 
A further next step is to consider other types of machine learning tasks such as regression, 
or unsupervised tasks such as clustering.
 More fundamentally, the PS score 
is only one approach to model root causes of errors, and it is well worth exploring other 
models and approaches. }

\bibliographystyle{ieeetr}
\bibliography{privacy,ml,causality}

\clearpage
\appendix
\section{Appendix}\label{sec:appendix-ml-algs}
In this section we present the details of two popular Machine Learning algorithms, 
Logistic Regression (Section~\ref{sec:ml-logreg}) and 
Gradient Boosted Decision Trees (Section~\ref{sec:ml-decision-trees}) as a reference to the reader.
The details are standard and can be found in any Machine Learning textbook~\cite{bishop}.
We focus on deriving the constraints used in the approximate models of the algorithms found in the main portion of the paper (Sections~\ref{sec:lr} and \ref{sec:dt}).

\subsection{Logistic Regression (LR)}\label{sec:ml-logreg}
Logistic Regression (LR) is a popular statistical classification model based on the sigmoid
scoring function:

\[h(Z) = \frac{1}{1+e^{-Z}},\]
where $Z = \theta^T\cdot x$ with an $m$-dimensional feature vector $x$ for a sample instance $\gamma$,
and an $m$-dimensional vector $\theta$ of inferred classification parameters.
In this work, we consider LR applied to the binary classification problem in which
instances with score $h(\theta^T\cdot x) < 0.5$ are classified $c(x)= -1$ 
and those with score $h(\theta^T\cdot x) \geq 0.5$ as $c(x) = 1$.

Learning a classifier is posed as an optimization problem over the space of $\theta$ values. Specifically,
the LR's algorithm attempts to find values of $\theta$ that maximize the log-likelihood function:

    \[L(\theta) = \sum_{(x^l,y^l) \in T}^N\log h(y^l z^l),\] where $z^l = \theta^T \cdot x^l$.

The problem then explores a convex landscape of $\theta$ vectors, and an iterative
 procedure for finding $\theta$ candidates (at some step $K$) can be given using the following recursive equation:
 
    \[\theta^K = \theta^{K-1} + \alpha\cdot\nabla L^{K-1},\]
where $\alpha$ is a step-size in $m$-dimensional space and the $i$-th component of the gradient $\nabla L^{K-1}$ is:

    \[ \nabla L_i^{K-1} = \sum_{(x^l,y^l) \in T}y^l x^l_i h(-y^l z^l)\]
with $z^l = (\theta^{K-1})^T \cdot x^l$ for iterates.

Viewing the above equation as a structural equation, we see every training point contributes to the gradient computation.
This is a problem as it implies a causal dependence from every input training point to the gradient computed at every iteration.
Additionally, in practice a complex optimization algorithm such as Stochastic Gradient Descent (or a variant)
is used to explore the search space.

Unfortunately, as we already stated in Section~\ref{sec:typical-scenario} computing the classifier weights $\theta$ involves recursive constraints in both the features and labels of 
\emph{all training instances} (details are found in Section~\ref{sec:ml-logreg}). 
These problems make causal inference infeasible even for state-of-the-art statistical inference techniques.
To reduce the complexity of constraints relating training samples to misclassifications $\varphi(c)$
our approximate causal model $\widehat{\cal{A}}$ for Logistic Regression uses profiling information
from the reference library implementation $\cal{A}$ of logistic regression. 

\thetech performs a single run of $\cal{A}$ on $\Gamma$ to records the values of the parameters essential 
to computing the weights $\theta$ inside the sigmoid classifier $h$: 
iterates of $\theta$ and line search steps $\alpha$
in the stochastic gradient descent variant of the search algorithm.
These parameters (denoted by hats) are then used to simplify key constraints relating training points to 
the internal computations resulting in misclassifications such as:

\begin{align}
\theta^K = \widehat{\theta}^{K-1} + \widehat{\alpha}\nabla \widehat{L},
\label{Equation:theta-K2}
\end{align}
where $\widehat{\theta}^{K-1}$ is the $(K-1)$-th iterate computed by the library implementation, and
$\widehat{\alpha}_i$ is the step size the implementation took in the along the $i$-component of the gradient:

\[ \nabla \widehat{L_i} = \sum_{(x^l,y^l) \in T}Y^l x^l_i h(y^l \widehat{z}^l).\]
Here $\widehat{z}^l = (\widehat{\theta}^{K-1})^T \cdot x^l$ includes the profiling constant.
Additionally, we memoize classification
scores $h(\theta^Tx)$ and reuse across computations for different training labels $Y^i$ using the following
well-known property of the sigmoid function $h(-Z) = 1 - h(Z)$.

Because the simplification in~\eqref{Equation:theta-K2} allows to compute any $\theta$ iterate
(say, the last one, or iterates upto a fixed depth $d$) by treating previous iterates as constants,
we can write misclassifications constraints in a simple linear form:

\begin{align}
\varphi(c)\ :\ &h_{\theta}(x) \geq \frac{1}{2}  \Longleftrightarrow \theta^T x \geq 0,\text{ and}\label{Equation:PhiGeq0-2}\\
&h_{\theta}(x) < \frac{1}{2} \Longleftrightarrow \theta^T x < 0,\label{Equation:PhiLeq0-2}
\end{align}
where $\theta$ is the final value that the library implementation would compute given 
the previous iterate $\widehat{\theta}^{K-1}$.
The simplification translates $\varphi(c)$ directly into a linear constraint over the labels of training instances:

\begin{align*}
\eqref{Equation:PhiGeq0-2} \Leftrightarrow \sum_{i=1}^{m}\left(\widehat{\theta}^{K-1}_ix_i - \widehat{\alpha}_ix_i \sum_{(X^l,Y^l) \in T}Y^l x^l_i h(y^l \widehat{z}^l)\right) \geq 0,\\
\eqref{Equation:PhiLeq0-2} \Leftrightarrow \sum_{i=1}^{m}\left(\widehat{\theta}^{K-1}_ix_i - \widehat{\alpha}_ix_i \sum_{(X^l,Y^l) \in T}Y^l x^l_i h(y^l \widehat{z}^l)\right) < 0,
\end{align*}
where feature vectors $X^l = x^l$, step size $\widehat{\alpha}$ and value of $h(y^l\widehat{z}^l)$ are known constants.

\subsection{Gradient Boosted Decision Trees}\label{sec:ml-decision-trees}
Decision trees are tree-shaped statistical classification models in which each internal node
represents a decision split on a feature value; and, associated with each leaf node is a
score $s$. The score $s_p$ of a particular point $p$ in the feature space $\mathbb{R}^n$ is computed by evaluating the
decision at each internal node and taking the corresponding branches until a leaf is reached.
The leaf score is then returned as score $s_p$.

Each leaf can be viewed as a region $R$ of $\mathbb{R}^n$ and a tree as a
partitioning $\{R_1, \cdots, R_L\}$ of the feature space $\mathbb{R}^n$.
Formally, the tree represents a piecewise constant function $h(x) = \sum_k s_k I(x \in R_k)$.
Where, $I(.)$ is the 0-1 indicator function and $s_j$ is the score
corresponding to region $R_j$.

Decision trees are learnt by recursively partitioning the training data on a
feature value that maximizes some statistic such as \emph{information gain} until either a
fixed number of points are left to be classified or a maximum tree height is reached.

Single decision trees are known to be prone to overfitting since with a large
enough tree, the model learnt can be very specific to the training data and
often each point can be classified correctly. Overfitting is countered by
learning a collection of small trees, known as an \emph{ensemble}, and
combining the scores associated with each tree in the ensemble.

One popular technique for learning an ensemble of decision trees is
\emph{gradient boosting}~\cite{Friedman00}. In gradient boosting, a simple tree
is initially learnt on the training data $T$. The residual error $E_{0}$ of
classifications is used as the new target and another tree is build over the
training set fit to the residual errors. This gives an iterative process of
refining the training errors that is usually bounded by a fixed number of
iterations $I_{\max}$.

As with the logistic regression classifier learning decision trees is framed as a loss minimization problem.
A loss function estimates the error on a training point for a given classifier.
For a loss function $L$, gradient descent proceeds as follows:
the initial tree $h_0(x)$ is set to a single partition with some constant score, usually $0$;
after $n-1$ trees have been learnt, resulting in a classifier $F_{n-1}(x)$, the next tree $h_{n}(x)$ is
fit to the gradient of the loss function at each training point:
$-\frac{\partial L}{\partial F_{n-1}}(x_i)$. A popular choice for a loss function
for binary classification is:
\[L(y, F_n(x)) = \log(1+e^{-2y\sigma F_n(x)}).\]
Here, $\sigma$ is a fixed constant known as learning rate.
The gradients with respect to this loss function is:
\[\bar{y}_{ni} = - \frac{\partial L(y_i, F_{n-1}(x_i))}{\partial F_{n-1}} = \frac{2y_i\sigma}{1+e^{-2y\sigma F_{n-1}(x)}}\]

A tree $h_n(x)$ is learnt with $\bar{y}_{ni}$ as the targets. The full details of the derivation
can be found in~\cite{Burges10}; however, the score
of the $k$th region (leaf) in $h_n$ is as follows:
\[s_{nk} = \frac{\sum_{x_i\in R_{nk}}\bar{y}_{ni}}
       {   \sum_{x_i\in R_{nk}}|\bar{y}_{ni}|(2\sigma - |\bar{y}_{ni}|)}\]

The tree learnt is $h_n(x) = s_{nk}I(x\in R_{nk})$ and the tree is added to the
existing classifier to obtain a new classifier $F_n(x) = F_{n-1}(x) + h_n(x)$.

The overall score is the sum of scores from each tree. Let $L$ be the total number of trees
and $K_n$ be the number of leaves in the $n^{th}$ tree, then the total score for a test
point can be written as:
\[
  s(x) = \sum_{n = 1}^L \sum_{k = 1}^{K_n} I(x\in R_{nk})s_{nk}
\]

If $s \geq 0$ the point is classified as $1$ else $-1$.

We now describe our model for approximating the effect of small perturbations
in the input training labels for the library implementation of Gradient Boosted
Decision Trees. Here we employ the robustness assumption to simplify the setting.
We assume that the trees do not change with
small perturbations in the input training labels. Now estimate the change in
the score of a leaf $s_{nk}$ due to a flip of label $y_i$ to $-y_i$. For the purposes of estimating the
scores for the tree $h_n(x)$ we assume that $F_{n-1}(x)$ does not change due to
the flip.  Therefore, the new value of the gradient is:
\[\bar{y}_{ni}' = \frac{-2y_i\sigma}{1+e^{2y\sigma F_{n-1}(x)}}
                = -2y_i\sigma(1 - \frac{\bar{y}_{ni}}{2y_i\sigma})\]

Also we note that $|\bar{y}_{ni}|(2\sigma - |\bar{y}_{ni}|)$ remains unchanged
under the same assumptions. Therefore, the new score $s'_{nk}$ due to flipping
$y_i$ to $-y_i$ is:
\begin{align*}
s'_{nk}
 &= s_{nk} + \frac{I(x_i\in R_{nk})(\bar{y}_{ni}' - \bar{y}_{ni})}
                {\sum_{x_i\in R_{nk}}|\bar{y}_{ni}|(2\sigma - |\bar{y}_{ni}|)}\\
 &= s_{nk} + \frac{-2 y_i \sigma I(x_i\in R_{nk})}
                {\sum_{x_i\in R_{nk}}|\bar{y}_{ni}|(2\sigma - |\bar{y}_{ni}|)}
\end{align*}

We calculate the score updates for each of the leaves in each tree to calculate
the total change in score.  Therefore the overall score $s'$ for an assignment
$\vec{Y}$ to the training labels is:

\[
  s'(x) = s(x) + \sum_{n = 1}^L \sum_{k = 1}^{K_n} \sum_{i = 1}^N \frac{(Y_i - y_i) \sigma I(x_i\in R_{nk})I(x\in R_{nk})}
                {\sum_{x_i\in R_{nk}}|\bar{y}_{ni}|(2\sigma - |\bar{y}_{ni}|)}
\]

The condition $\varphi(c)$ which is equivalent to $s \geq 0$ or $s < 0$ is
therefore a linear constraint on $Y_i$'s. This is the result we take away from 
this section and use in Section~\ref{sec:dt}.

\end{document}